# Text Annotation Handbook
## A Practical Guide for Machine Learning Projects

This handbook is a hands-on guide on how to approach text annotation tasks. It provides a gentle introduction to the topic, an overview of theoretical concepts as well as practical advice. The topics covered are mostly technical, but business, ethical and regulatory issues are also touched upon. The focus lies on readability and conciseness rather than completeness and scientific rigor. Experience with annotation and knowledge of machine learning are useful but not required. The document may serve as a primer or reference book for a wide range of professions such as team leaders, project managers, IT architects, software developers and machine learning engineers.

## Main Editor

Felix Stollenwerk (AI Sweden)

## Contributors[1]

Felix Stollenwerk (AI Sweden)
Joey Öhman (AI Sweden)
Danila Petrelli (AI Sweden)
Emma Wallerö (Ekonomistyrningsverket)
Fredrik Olsson (Gavagai)
Camilla Bengtsson (Gavagai)
Andreas Horndahl (Arbetsförmedlingen)
Gabriela Zarzar Gandler (King)

## Acknowledgements

This work is a result of the "Databeredskapsverkstad" (Data Readiness Lab) project funded by Vinnova (Sweden's innovation agency) under grant 2021-03630.

---

[1] Contributors have participated in discussions and text reviewing. Authors are specified at the beginning of each section.

# Table of Contents

# 1 Introduction

*written by Felix Stollenwerk*

Many organizations have substantial amounts of text data stored in their internal systems. This data often provides a wide range of opportunities, such as a deeper understanding of the organization's clients and domain, as well as automated and improved services through the application of machine learning. However, oftentimes, the data first needs to be annotated in order to provide value or leverage its full potential. Annotation means that whole documents, sentences or words are assigned certain labels. Two examples—a fictitious use case where user reviews are categorized, and a real-world use case where key information in job ads is extracted [1] —are shown in **Fig. 1**.

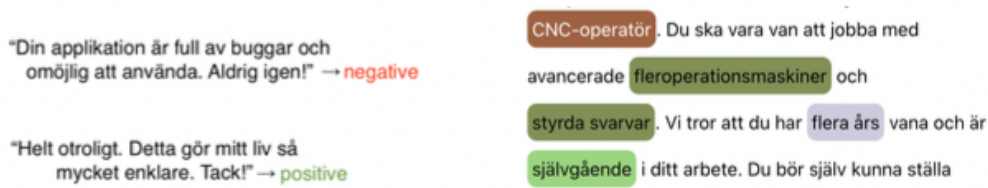

**Figure 1:** Two examples of annotated text. On the left, we show user reviews that are assigned the label "positive", "neutral" or "negative". This corresponds to annotation on the document level. On the right, an excerpt of an annotated job ad is shown. Here, the annotation takes place on the word level. The labels shown are called "job title" (brown), "hard skill" (dark green), "soft skill" (light green) and "experience time" (light violet), see [1] for more details.

To understand why annotation can be useful, let's consider the user review use case example from **Fig. 1**. Say your organization receives a lot of reviews every day, which at present are read and analyzed manually. Your aim is to create a system that screens incoming reviews and automatically assigns labels ("positive", "neutral" or "negative") to them. This task can be solved by supervised machine learning (see **App. A** for an introduction to the topic). In order for a machine learning model to learn how to solve the problem, however, one needs to create some training data that it can learn from. In our example, this would be a set of reviews together with their assigned labels.

These training datasets usually need to be manually created and this often needs to be done in-house, for instance because the annotation requires domain-expert knowledge or the data contains sensitive information. The annotation poses a challenge for many organizations as it can be a tedious and expensive process. It is in fact often a major bottleneck that prevents organizations from successfully applying machine learning and transforming it into a useful, real-world product.

That is why we wrote this handbook. Its purpose is to help you and your organization tackle the challenge of annotation. We aim to provide a useful how-to guide that enables you to approach the problem of annotation efficiently and take the right decisions from the start. It begins with a gentle and easy-to-digest introduction to the most important concepts, while more technical and special topics are covered in later sections. More specifically, it first outlines an essential principle that we think should underlie any work on annotation (**Sec. 2**), followed by a presentation of useful methods (**Sec. 3-5**) and helpful annotation tools (**Sec. 6**). Subsequently, common pitfalls that should be avoided are discussed (**Sec. 7**). The handbook concludes with considerations from a business perspective (**Sec. 8**) and a discussion of ethical issues and regulations (**Sec. 9**).

# 2 Annotation Principle

*written by Felix Stollenwerk*

Annotation tasks can vary significantly in complexity. To illustrate this, let's return to the two examples mentioned in the introduction and shown in **Fig. 1**: User review sentiment analysis and annotated job ads.

The former poses a rather simple problem. A human can usually figure out the correct sentiment easily and in no time. Hence, the task can be accomplished quickly and the assigned labels will be correct in the vast majority of cases.

In contrast, the example of annotated job ads is a lot more challenging. One issue among many is to choose the right boundaries for an entity. For instance, in **Fig. 1**, should the skill required by a job seeker be "fleroperationsmaskiner" or "avancerade fleroperationsmaskiner"? An annotator will need time to think about such things, and two annotators will often come up with different results. This may easily lead to insufficient amounts of data as well as poor data quality. Both are problems that may affect the performance of the resulting machine learning model negatively.

Fortunately, there are methods to mitigate those issues. However, the methods add technical complexity and pose an overhead that is unnecessary and obstructive for simple annotation problems like the user review sentiment analysis example. Employing those methods is a trade-off and should be considered carefully. This leads us to a very essential principle, visualized in **Fig. 2**: **The employed methods should match the difficulty of the annotation task.**

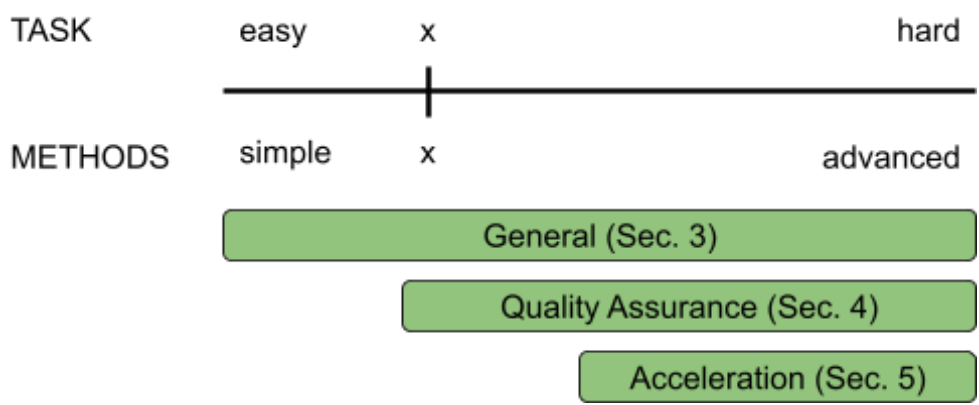

**Figure 2:** Spectrum of difficulty for an annotation task. The employed methods should be chosen such that they match the difficulty of the task.

There are universal methods that should always be used, irrespective of the nature of the annotation task. Those will be covered in **Sec. 3**. All but very simple problems usually require quality assurance measures, which are described in **Sec. 4**. Especially for very challenging problems, the acceleration methods discussed in **Sec. 5** may be useful.

# 3 Annotation Basics

*written by Felix Stollenwerk & Emma Wallerö*

In this section, we describe initial steps that we consider indispensable for any annotation project. These are a proper definition and understanding of the task (**Sec. 3.1**) and the creation of annotation guidelines (**Sec. 3.2**). Afterwards, **Sec. 3.3** discusses a simple annotation workflow which may serve as a guideline.

## 3.1 Understanding of the Task

Annotation means taking some sort of text (input) and assigning some sort of label (output) to it. However, the type of input and output can vary for different annotation tasks. Before starting the annotation process, one should have a basic understanding of the task at hand. In addition to the input and output, this also includes knowledge of how the performance of a model that is trained on the task can be assessed.

**Input text**
Labels can be assigned to input text on different levels. The most common levels are whole documents, paragraphs, sentences and words. In **Fig. 1**, we have seen two examples that assign labels to documents and words, respectively. Also pairs of the aforementioned structures can be annotated. For instance, two sentences can be assigned a label based on how similar they are ("semantic textual similarity"). This is illustrated in **Fig. 3**.

|   | sentence1 | sentence2 | similarity_score |
|---|---|---|---|
| 0 | A girl is styling her hair. | A girl is brushing her hair. | 2.5 |
| 1 | A group of men play soccer on the beach. | A group of boys are playing soccer on the beach. | 3.6 |

**Figure 3:** Examples of two pairs of sentences and their assigned semantic textual similarity score. The score is a numerical value between 0 (least similar) and 5 (most similar). The figure is taken from [2].

**Output labels**
There are two kinds of output labels. Some are categorical (e.g. "positive" vs. "negative"), some are numeric (e.g. "1.4" vs. "2.3"). The tasks to predict the former are called "classification" tasks, while the tasks to predict the latter are referred to as "regression" tasks. The two examples from **Fig. 1** are both classification tasks, while semantic textual similarity is a regression task as the similarity is usually specified as a number. Classification tasks are characterized by their labels, and the number of labels may vary. In contrast, regression tasks are characterized by their output range, which is often an interval between two integer numbers.

**Tasks**

Tasks are essentially defined by their input and output. In **Tab. 1**, we show a few examples of tasks and their features.

| Task | Task Group | Input | Output | Metrics | Example |
|---|---|---|---|---|---|
| Sentiment Analysis | Text Classification | Document | Classes | precision, recall, f1 | see Fig. 1 |
| Named Entity Recognition | Token Classification | Word | Classes | precision, recall, f1 (entity) | see Fig. 1 |
| Semantic Textual Similarity | Sentence Similarity | Sentences | Numeric | Pearson or Spearman Correlation | see Fig. 3 |

**Table 1:** Non-exhaustive list of annotation tasks and their characteristics. For a complete overview of tasks and details about the listed ones, see for instance https://huggingface.co/tasks.

**Assessment of model performance**

After a model is trained on the annotated data, one is interested in how well it has learned to perform the task. The model's performance can be assessed with so-called metrics. A metric compares the model's predictions with labels from the manually annotated data and measures how well they agree. An example is shown in **Fig. 4**.

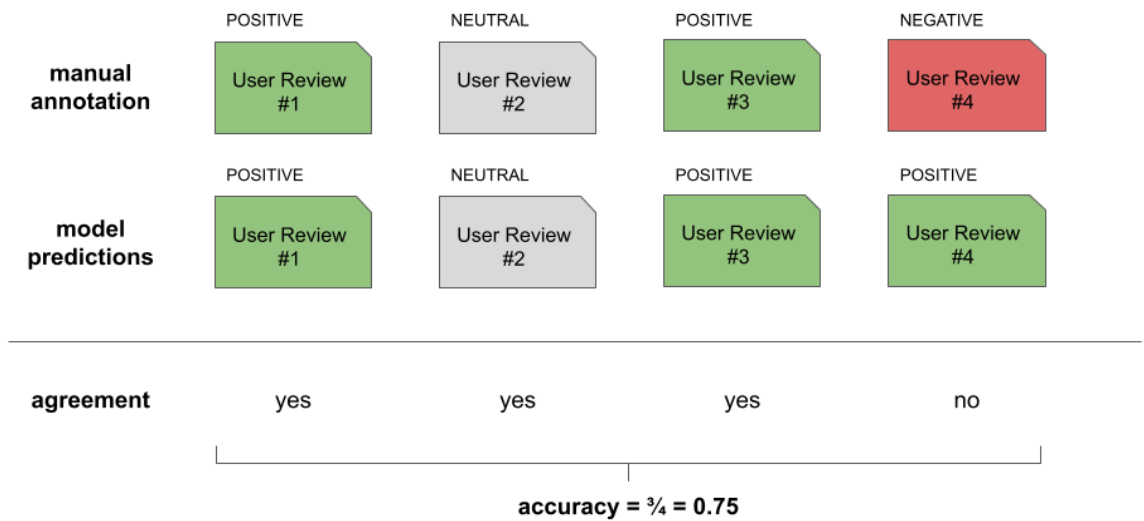

**Figure 4:** Example of 4 user reviews that are assigned one of the labels "positive", "neutral" and "negative" each. The manual annotations are compared to the model predictions and a metric (in this case: accuracy) which measures the performance of the model is computed.

Since agreement can be measured in different ways, there are usually multiple metrics to choose from. However, a given task usually restricts the number of metrics that are suitable to assess the model's performance to one or at least very few. General metrics that can be considered for classification tasks are accuracy, precision, recall and f1 score, while common metrics for regression tasks are for instance RMSE or Pearson/Spearman correlation. For a general introduction to those metrics, see e.g. [3], [4]. Some tasks, like Named Entity Recognition, have special task-specific metrics, see e.g. https://huggingface.co/docs/evaluate/choosing_a_metric.

It is important to note that the model performance is to be evaluated on a so-called test dataset that the model has not seen during training. To this end, the available annotated data is usually divided into a training and a test dataset, see **App. A** for more details.

**Every use case is unique**

Once you have understood the general type of your task and the metrics that allow you to measure the quality of model predictions, it is time to have a closer look at the characteristics of your specific task. Let's return to the user review sentiment analysis example from **Fig. 1**. While it is a text classification task that takes input sentences and assigns categorical output labels, the number and very definition of the labels are often unique for a given use case. It is important to clearly define what the labels represent, both with regard to the annotation process (it will help the annotators to provide consistent, high-quality annotations) and the use of the model in applications later on (it will help users to understand what exactly the model predictions imply).

An essential helper tool to ensure that the annotation task is well-defined are annotation guidelines, which will be covered in the next part.

## 3.2 Creation of Annotation Guidelines

Annotation guidelines are critical to an annotation project since it is through these that you define labels, features and annotation units. By formulating the instructions you are ensuring a certain degree of reproducibility of the annotation project.

Simple annotation guidelines should contain at least:
1. A task description
2. A label description
3. Some annotation examples

A template for simple annotation guidelines can be found in **App. B**.

The annotation guidelines should not be more complicated than necessary. It is important that the guidelines are adapted to the annotators. If you wish to involve many annotators with varying knowledge concerning your task it is necessary that you present adequate background information. Clearly defined units and labels are important in order for annotators to assign the data units to the same labels as one another.

When designing the guidelines it may be helpful to discuss the material and task with the annotators and try to define some data together. By doing this, some uncertainties that you had not thought about yourself might come up. These problems can then be defined and you can explain how to handle similar situations in the guidelines.

**Use case example**: An annotator is not sure how to handle a review that contains both positive and negative sentiment. The review follows: "My brother thought this dishwasher is the best ever. I don't agree". How should such a review be annotated? Since the opinion of the reviewer is negative, it could be argued that this sentiment should be preferred. However, taking into account the ambiguity of the review, one could also justify giving it a "neutral" label.

By putting the guidelines in a dedicated document, you also easily make sure that all annotators receive the exact same instructions. This will encourage the annotators to produce more consistent annotations. It is important to have well defined guidelines as well as a good understanding of the guidelines because these things will affect the quality of your data, which in turn will most likely affect the performance of your language model.

## 3.3 Basic Annotation Workflow

Once the annotation guidelines are set up, one may in principle begin the annotation process. The simplest annotation workflow one can imagine is the following: A dataset is annotated by a single person in one go. Subsequently, a model is trained and evaluated on the annotated data. Although this workflow, which we will refer to as "simplistic workflow" might seem appealing, it is hardly useful in practice as it entails a few problems.

Firstly, it is virtually impossible to know beforehand how much data one should annotate. To understand why this is the case, take a look at **Fig. 5**. It shows a so-called learning curve of a model, i.e. the dependency of the model performance on the size of the dataset. Typically, in the beginning, the model performance improves rapidly as the dataset size is increased. However, the gain from more data becomes less and less, and eventually, a plateau is reached where additional data has little or no effect anymore. In the example of **Fig. 5**, the beginning of the plateau is located somewhere around dataset size = 2000. This is usually the "sweet spot" that one aims to reach: One annotates as much data as needed to get near the best possible model performance, but not more, as this probably isn't worth the effort.

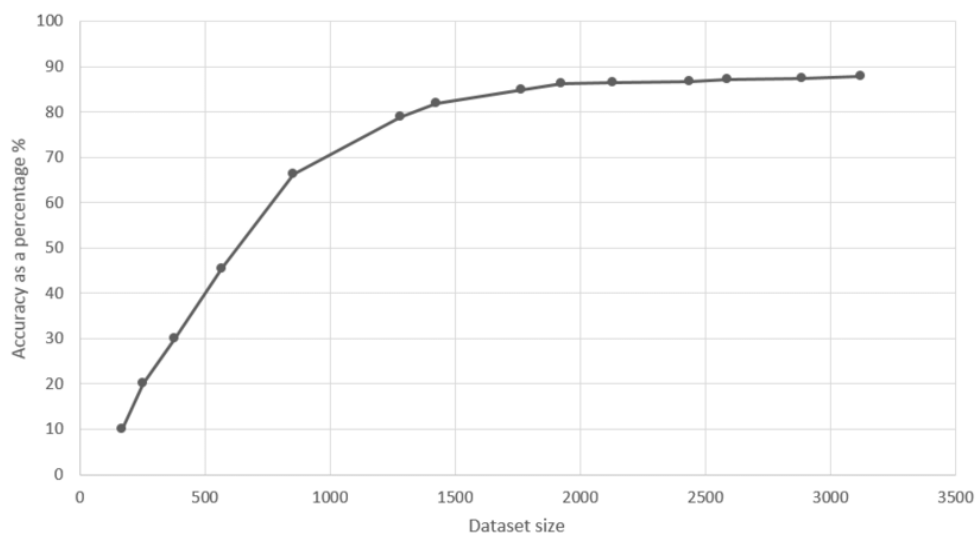

**Figure 5:** Learning curve which exemplarily shows the performance of the model (here: accuracy) on the test dataset as a function of the size of the dataset that the model was trained on. The figure is taken from [5].

The problem with the aforementioned simplistic workflow is that the dataset size is predefined. After a model is trained, one ends up with only a single point on the otherwise unknown learning curve, which makes it impossible to tell whether one is below or above the beginning of the plateau. In that case, it is likely that one either has too little or too much data, resulting either in poor model performance or a waste of resources. A relatively simple way to avoid this issue and make sure to annotate the right amount of data is to **annotate iteratively in batches** and **track the model performance after each iteration**, as illustrated in **Fig. 6**.

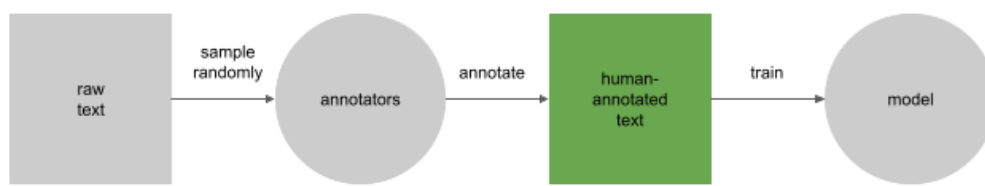

**Figure 6:** One iteration in the basic annotation workflow. A reasonable batch size (in terms of documents, sentences or words) is defined. Then, at each iteration, a batch of unlabelled text is randomly chosen and annotated. Subsequently, a model is trained on all the available annotated data, i.e. the data annotated in the present and all previous iterations.

This way, one is able to monitor the development of the model performance similar to what is shown in **Fig. 5**. Once the model performance stops improving (or is sufficiently good), the annotation process is stopped. Note that in order for the model performance after each iteration to be comparable, the model needs to be evaluated on the very same test dataset every time. Hence, it is advisable to first create a carefully curated test dataset, and then use the iteratively annotated data as training data[2].

A second problem with the outlined simplistic workflow is that it involves a single annotator. It is often way too much effort for a single person to annotate all the required data alone. Therefore, we proceed on the assumption that multiple annotators share the task. One way to do this is to split each batch between the annotators as shown in **Fig. 7**.

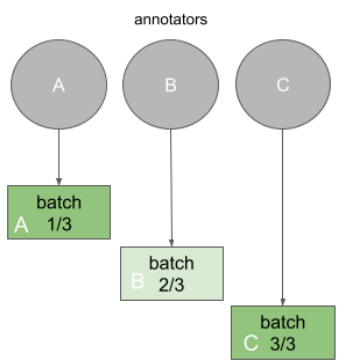

**Figure 7:** Illustration of a batch of data being split between 3 annotators (A, B, C) in the iterative process. Each annotator gets assigned a third of the batch. Dark (light) green indicates good (poor) annotation quality. The figure is taken from [1] and has been slightly adjusted.

---

[2] There are a few caveats here to consider in practice. We will discuss them in **Sec. 7**.

The iterative workflow illustrated in **Fig. 6** and **Fig. 7** should suffice for very simple annotation tasks (cf. **Fig. 2**). In addition, it will serve as the basis for more advanced workflows to be discussed in the following sections (**Sec. 4** and **Sec. 5**).

# 4 Quality Assurance

*written by Felix Stollenwerk*

Annotations done by humans are never deterministic, but subject to natural variation. Two annotators may have a different understanding of a given text or the annotation guidelines and thus provide different annotations. Even one and the same person might annotate the same text differently on two different days, based on external factors (e.g. mood, tiredness, distraction) that bring in randomness. Inconsistent or incorrect annotations are not necessarily a problem if they appear rarely, which is typically the case for simple annotation problems. However, if they occur too often, they can impair the quality of the data to such an extent that the performance of the resulting model suffers.

In those cases, the annotation part of the basic workflow described in the previous section can be modified in order to ensure high quality data, see **Fig. 8**.

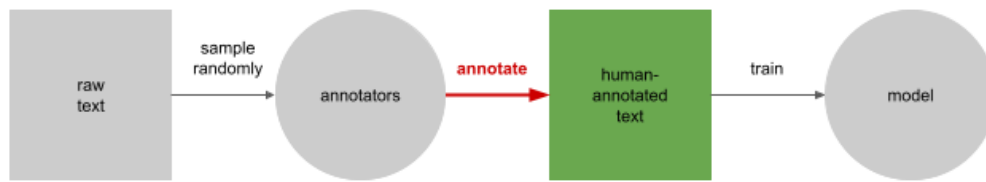

**Figure 8:** One iteration in the basic annotation workflow with quality assurance (highlighted in red).

Instead of each annotator working exclusively on their own set of documents (as shown in **Fig. 7**), we introduce some kind of interaction between the annotators.

## 4.1 Reviewing

A very simple way to do this is reviewing, where annotations by one annotator are double-checked by another annotator, see **Fig. 9**.

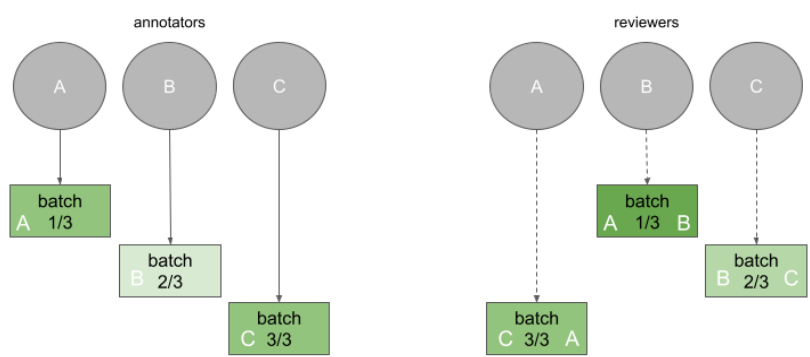

**Figure 9:** A simple reviewing process. First, each person annotates their part of the batch. Subsequently, the parts are permuted and assigned to other annotators, who review the annotations.

Different scenarios for the reviewing process itself are possible. The reviewer could simply correct the annotations without consulting the original annotator. However, it is advisable that the reviewer and original annotator discuss the cases where they disagree. Not only will this likely lead to even better data quality, it also helps the annotators to gain experience and get better at annotating.

## 4.2 Cross-annotation

Another, more advanced way to assure high quality data is cross-annotation. This means that each document gets annotated multiple times in the first place, by different annotators. The existing, individual annotations may then be compared, and the amount to which they coincide can be quantified by computing the so-called inter-annotator agreement, see **App. C** for details. A high inter-annotator agreement serves as a proxy for high data quality.

Now that we are able to measure data quality, how do we go about improving it? And which of the existing annotations per document shall we use? The answer to those questions is: Resolve the conflicts between the individual annotations. The annotation quality after conflicts are resolved will likely improve since annotations will become more consistent.

There are several ways to cross-annotate and resolve conflicts in practice. One particular algorithm, illustrated in **Fig. 10**, has been described in [1] and implemented in [6]. In essence, the batch of data that is to be annotated in a single iteration is divided into as many parts as there are annotators. Subsequently, each part is assigned to two different annotators in such a way that each annotator gets the same workload. The annotators then work on their tasks individually. The two individual annotations per part are merged and conflicts are highlighted. An example of this procedure for the use case of annotated job ads (cf. **Fig. 1)** is shown in **Fig. 11**. Finally, all annotators go through the merged annotations and resolve conflicts together. This procedure is efficient in the sense that the majority of annotations are handled by only two annotators, while merely the conflicts have to be taken care of by more people. An additional benefit is that the group gets the chance to discuss difficult cases together, a learning experience which may help to create more consistent annotations in future iterations.

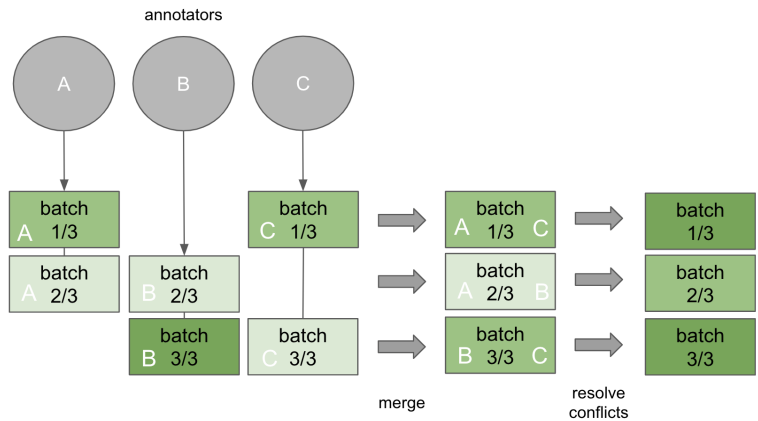

**Figure 10:** Annotation strategy which involves cross-annotation. Each sample is annotated by two different annotators. After merging, the conflicts are discussed and resolved in a collective effort. The figure is taken from [1]. Compare to **Fig. 7**.

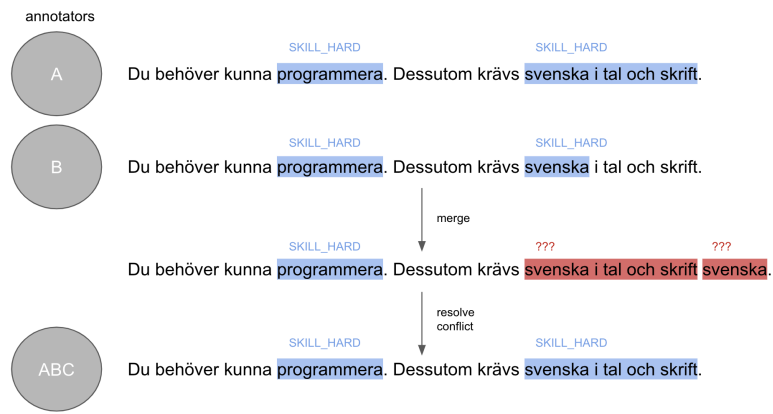

**Figure 11:** Excerpt of a job ad that gets annotated separately by two annotators A and B. The result is merged before the conflicts are resolved together by all annotators. The figure is taken from [1].

# 5 Acceleration

*written by Felix Stollenwerk & Gabriela Zarzar Gandler*

Annotation is usually time and labor intensive. The measures for quality assurance described in the previous section even pose an additional effort. In this section, we will discuss some methods that may help to speed up the annotation process.

There are essentially two ways to do this:
    A.  Reduce the effort to annotate a given amount of data
    B.  Reduce the amount of data that is needed to reach the model performance plateau (cf. **Fig. 5**)

## 5.1 Bootstrapping

Bootstrapping is an acceleration method of type A. As mentioned in previous sections, our iterative approach involves the repeated training of a model after each step. Naturally, with more training data, the model predictions become increasingly more accurate, see **Fig. 5**. The idea behind bootstrapping is to leverage the model predictions to help the annotators. Instead of making the annotators work on raw text data, one lets the model annotate the data before it is presented to the annotators, see **Fig. 12**.

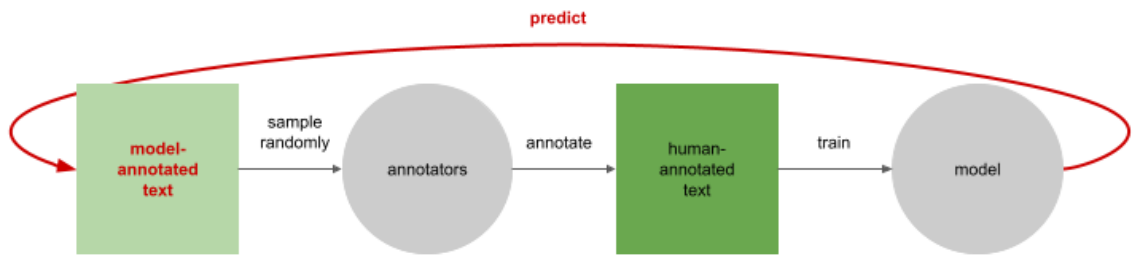

**Figure 12:** One iteration in the annotation workflow with bootstrapping (highlighted in red). Compare to **Fig. 6**. Bootstrapping introduces a feedback loop between the iterations. Note that the light green color for the model-annotated text symbolizes medium annotation quality.

This way, the annotators will "only" have to correct the model predictions, instead of doing the annotation from scratch. Usually, this reduces the effort significantly, especially in later stages of the iterative process as the model performance gets better. We refer to [1] for more details and things to take into consideration when using bootstrapping.

## 5.2 Weak Supervision

Weak supervision is also an acceleration method of type A. As with bootstrapping, the raw data is pre-annotated before it is presented to the annotators. However, the pre-annotations do not stem from a model trained on human-annotated data, but from so-called noisy programmatic labelers. This essentially means that domain experts manually define text patterns that are associated with labels. The raw text data is then screened for those text patterns and assigned the corresponding labels.

Let's consider the two examples from the introduction (**Sec. 1**) to see what this could look like in practice. For sentiment analysis, the user reviews could be screened for words like "good" and "excellent", which are associated with the label "positive". In contrast, words like "disappointing" as well as "not" followed by a positive word are associated with the label "negative". For the use case of annotated job ads, a set of words that refer to skills could be automatically be searched for and assigned the label "hard skills" if found (cf. **Fig. 1**).

Note that annotations acquired by weak supervision are noisy and not of the same quality as annotations provided by humans. Also, in contrast to bootstrapping, they are static and do not improve over time. However, they can be generated with relatively little effort (assuming that one has basic programming skills and domain knowledge). Hence, they may be of help, especially in early stages of the iterative annotation process.

A code example that implements weak supervision for sentiment analysis can be found here: https://github.com/heartexlabs/label-studio-sdk/blob/master/examples/weak_supervision/weak_supervision.ipynb.

## 5.3 Active Learning

Active Learning is an acceleration method of type B. As discussed in previous sessions, annotating data samples can be time-consuming. In addition to that, in machine learning problems, all data samples are typically not equally informative. Hence, wouldn't it be advantageous to save annotation time by identifying the most informative data samples and annotating them only?

To achieve this, instead of asking annotators to go through randomly sampled data points, we can provide annotators with the most informative data points first. Here, "informativeness" is the quality of data points that are especially useful for the trained model to become better at the given task. Active learning can easily be incorporated in the iterative process described in earlier sections (see **Fig. 13**):
1. Given the current sets of annotated and non-annotated data, a learning algorithm detects the most informative data points out of the ones left non-annotated, which are then sampled and presented to the annotators
2. Annotators provide annotations to these data points
3. The learning algorithm then takes the newly annotated data into account and the procedure repeats from 1.

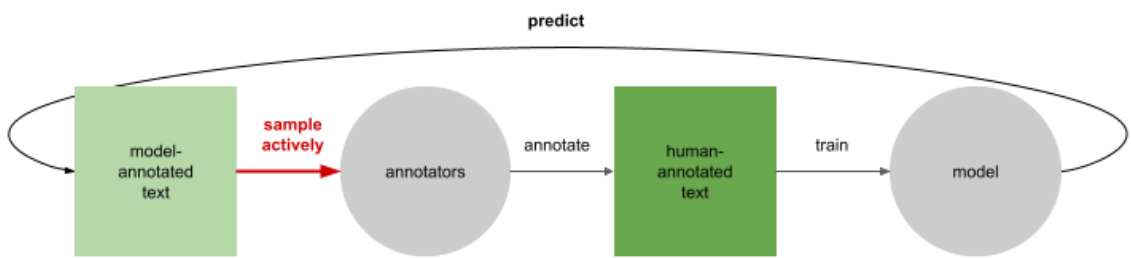

**Figure 13:** One iteration in the annotation workflow with active learning (highlighted in red). Compare to **Fig. 12**.

Since there is a learning algorithm behind the scenes strategically picking the most informative data points, the amount of annotated data needed to reach a model performance plateau (see **Fig. 5**) can be much smaller than the amount required if providing annotators with randomly sampled data points.

The described iterative process lies at the heart of the field of active learning. There are numerous variations of active learning that differ in the way the most informative data points are defined and chosen. See [7] for an overview of research works in this field.

# 6 Annotation Tools

*written by Felix Stollenwerk*

In the previous sections, we have discussed the *theoretical foundations* and some useful methods that should give you the knowledge to approach your annotation project efficiently. In this section, we cover the annotation tools that can be employed *in practice*.

There are plenty of annotation tools to choose from. We have compiled a table[3] which gives a detailed (yet non-exhaustive) overview of many important tools and their features. In particular, it tells you which of the methods described in **Sec. 4-5** are supported by which tool. An excerpt of the table is shown in **Fig. 14**, while the complete table is available online[4].

| | Tool | Prodigy | Label studio | Doccano |
|---|---|---|---|---|
| 1 | | | | |
| 4 | Why is this tool included in this overview? | One of the most popular annotation tools | One of the most popular annotation tools | One of the most popular annotation tools |
| 5 | Date of column update | 2023-04-13 | 2023-04-13 | 2023-04-13 |
| 6 | SaaS | No | No | No |
| 7 | Self hosted | Yes, local server with web interface for annotation. | Yes | Yes |
| 8 | Containerized | No, no official image but installation instructions cover Docker. Community solutions exist online | Yes, official Docker container available | Yes, official Docker container available |
| 9 | Reviewing (Sec. 4.1) | Yes | Yes | Yes |
| 10 | Cross-annotation (Sec. 4.2) | Yes | Yes, using an external library* | Yes, using an external library* |
| 11 | Bootstrapping / external model predictions (Sec. 5.1) | Yes, using spaCy models | Yes | No, but pre-annotated data can be imported |
| 12 | Weak supervision (Sec. 5.2) | No, not advertised but can be implemented and example use case(s) exist | Yes, advertised with example code | No, not advertised |
| 13 | Active Learning (Sec. 5.3) | Yes, NER, text classification, dependecy parsing, pos tagging | No, but it can probably be implemented. | No, but it can probably be implemented |

**Figure 14:** Excerpt of the annotation tool overview table. Note that only 3 annotation tools (columns) and 10 features (rows) are shown.

As explained previously (cf. **Sec. 2**), each use case has its own challenges and the annotation methods should be chosen accordingly. Consequently, the annotation tool that is best suited for a use case may vary as well. However, despite this, as an organization with multiple (even future) use cases, it might be better to stick to a single annotation tool that fits the general requirements best, and use it repeatedly instead of employing a different annotation tool for each individual use case. The reason is that each annotation tool poses a technical overhead. For instance, their user interfaces all work differently, and the annotators need to become familiar with them. In addition, the programmatic access to the data and the annotation tool differs between the tools, since many of them use custom data formats, APIs, SDKs and clients.

---

[3] Note that the table was initially created by Fredrik Olsson and Magnus Sahlgren [8]. We updated it in the course of writing this handbook.

[4] https://docs.google.com/spreadsheets/d/e/2PACX-1vQV52O0KIwQO24J9D-bEGgVFwMWabm12PDwJ1RMCLp4voxP8dHyKHGG5ncst25Du-NE9Yu-G8YDRXE3/pubhtml

# 7 Annotation in Practice

*written by Felix Stollenwerk & Emma Wallerö*

The annotation workflow with its different variants, described in **Sec. 2-5**, together with the discussion of available annotation tools in **Sec. 6**, has proven to be very applicable in our own experience ([1], [9]). However, it is important to be aware of certain problems and pitfalls that may occur in the midst of the iterative annotation process and require adjustments. In this section, we cover some of the most common issues (without claiming to be exhaustive), and explain how they can be avoided or mitigated.

## 7.1 Refinement of Guidelines

Even if the annotation guidelines are well thought through, you may come across surprises and detect issues you were not yet aware of. Such issues could regard the discovery of an unwanted overlap between two classes or realizing that you want to add or remove a class for some reason. At this stage, it might be a good idea to revise and improve your guidelines and class definitions. If you realize that you have to change the guidelines at a later stage in the process, you will have to correct all the previously annotated data according to the new changes. It is never completely unavoidable to go back and forth in the iterative process like this. However, one should at least minimize the risk of having to do so, by preparing the annotation guidelines as thoroughly as possible before starting the actual annotation process.

There are several ways to improve the guidelines of your task. You can improve them yourself, have someone else correct them, or preferably perform an iterative annotation setting where you and a group of annotators perform several annotation cycles and collectively improve the guidelines after each iteration. A more thorough description of how this can be achieved will be described in this section.

Firstly, formulate your annotation guidelines as well as possible. Quality-check these by annotating a data sample yourself, following the guidelines. Then ask yourself the following questions:
- Is the initial description of the problem clear and concise? *A basic understanding of the task is crucial for the annotators.*
- Are the classes well described? Was it easy to annotate data for each class given the instructions?
- Are all the classes present in your annotated data?
- Are the classes mutually exclusive, that is, are they defined in such a way that they do not overlap? *Having overlapping classes might mean that the task for which the annotation is to be carried out is not well-defined, and thus that the understanding of the end-usage of the learned model is not well-understood.*

If your answer to any of these questions was no, you should do your best to adjust your guidelines. Some possible measures to take could be:
- Improve your class descriptions and consider adding information that might be missing.
- Rearrange the class system by adding, removing or merging classes (cf. **Sec. 7.2**).

Secondly, you start your iterative annotation process. One iteration might look something like this:
1. Prepare a small amount of data for annotation
2. Have annotators try out the guidelines in a live annotation setting
3. Ask annotators for feedback on the guidelines
4. Discuss ambiguous cases. Do the guidelines cover these? If not, add directions to help the annotators know what to do when handling tough cases.
5. Prepare updated and (hopefully) improved guidelines.

When you believe the guidelines seem good enough for production work, similar iterations can be performed but the second step shall be performed differently:

2. Have annotators annotate the data separately.

This will give you the opportunity to measure inter-annotator agreement, see **App. C**. By taking these measurements, you get an indication of the quality of your annotation and your guidelines. Poor inter-annotator agreement may require further improvement of the guidelines as well as discussion between all annotators. So, if you are not satisfied with the agreement you can continue to perform more iterations. If you are happy with the agreement you can move on to production, using your annotation guidelines to create labeled datasets.

Some suggestions regarding certain steps in the iteration:
1. If the data format/extraction is not set in stone, it can be useful to discuss this in the group as well.
4. See this as an opportunity to also give the tough cases a concrete label. These can then be used as examples for the annotators to look back on when in doubt while annotating tough examples in the future. It can also then be used for your final data set, if the guidelines do not change dramatically.

Furthermore, you should be open to add and adjust any steps you deem necessary to the iterative process. Since annotation projects can be very different to one another, there is no single general solution for all tasks. Lastly, it is always a good idea to look beyond the project for help and inspiration. Perhaps a similar task has been done before. Existing guidelines and data sets might help you forward if you get stuck.

## 7.2 Class System Adjustments

Class system adjustments were introduced in [1] and are a way to deal with problematic classes during the iterative annotation process. A class can be considered problematic, for instance, if throughout the iterations

- the annotators continuously have difficult discussions about annotations that involve the class (at the cross-annotation group sessions where conflicts are resolved, cf. **Sec. 4.2**)
- the inter-annotator agreement (cf. **Sec. 4.2** and **App. C**) with respect to the class remains low and does not improve
- the model performance with respect to the class remains low and does not improve

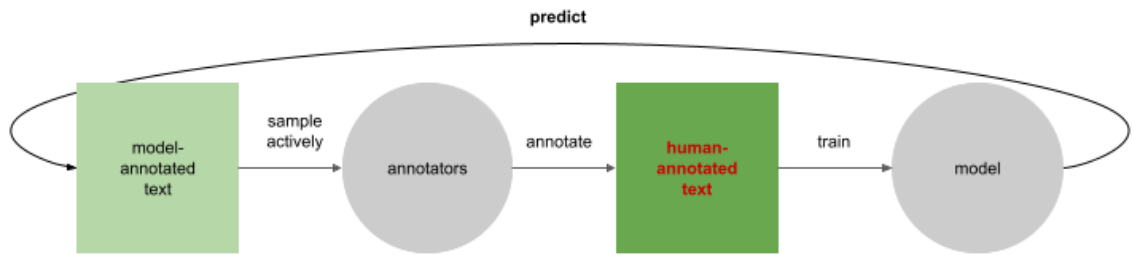

**Figure 15:** One iteration in the annotation workflow with class system adjustments (highlighted in red). Compare to **Fig. 6**.

A trivial way to deal with a problematic class is to just drop it. While this is very straightforward, it may not always be the best way to deal with the problem. Depending on the circumstances, it might be better to incorporate the class into another class or merge two problematic classes. The latter might for instance be useful if two classes get confused with each other by both the annotators and the model because they are too similar. However, there are certain caveats to take into account. Firstly, one has to consider whether the class incorporation or merge makes sense with respect to downstream applications of the model. Secondly, such operations require a refinement of the annotation guidelines (see **Sec. 7.1.**). Fortunately though, all three operations (drop, incorporate, merge) are deterministic transformations of the data that can be automated. Hence, the previously discussed revision of all previously annotated data is usually not necessary. We refer to [1] for details.

## 7.3 Representativeness of the Test Dataset

In our supervised machine learning setting, it is important that the test dataset accurately represents the real data distribution. If this is not the case, the model performance evaluated on the test dataset may not be a reliable predictor for the model performance applied to real-world data in production. This situation is illustrated in **Fig. 16**.

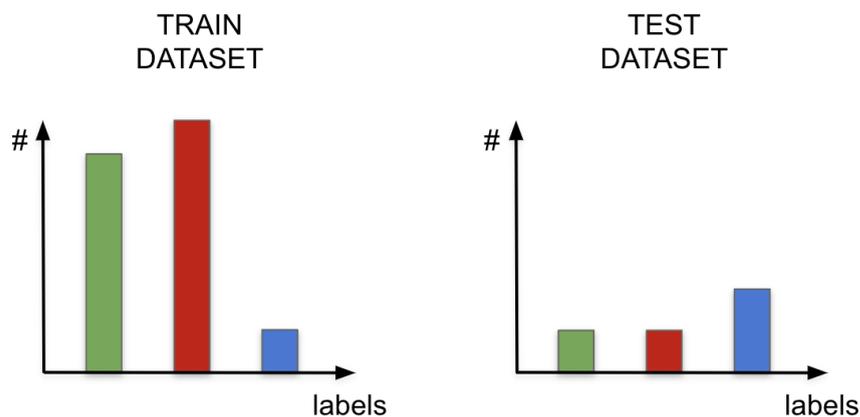

**Figure 16:** Undesirable situation where the representativeness of the test dataset is not given. The train dataset (left) is a larger and thus better sample of the real (unknown) label distribution. As one can see by comparing the two distributions, the green and red labels are underrepresented in the test dataset while the blue label is overrepresented. If a test dataset like this is used for model evaluation, the results may not be a good proxy for the model performance on real-world data.

As discussed in **Sec. 3**, the annotation workflow first includes the creation of a test dataset. At this point in time, however, one has no knowledge of the real data distribution and thus no means to assess whether the test dataset is a valid representation. Fortunately, there are two measures that can be taken to reduce the likelihood of having a test dataset that is not representative of the data distribution.

1. Large test dataset
   The size of the test dataset is typically relatively small compared to the training dataset. However, the larger the test set, the more likely it is that it resembles the real data distribution. Hence, it might be worth putting some extra effort into creating a slightly larger test dataset.

2. Monitoring
   As the training dataset is accumulated in an iterative manner **(cf. Sec. 3)**, one gains more and more information on how the data are distributed. One may for instance monitor the label distribution of the training and test dataset and compare them after each iteration.

If monitoring reveals that the differences between the datasets is too big, one may consider resplitting the data such that the two new dataset splits have more similar distributions. Note, however, that this entails that all previous model evaluations (for instance for monitoring of the model performance, cf. **Fig. 5**) need to be repeated as it wouldn't be fair to compare models evaluated on the old and new test dataset. Hence, resplitting the dataset should ideally be avoided in the first place by using a sufficiently large test dataset (see 1.).

# 8 The Business Perspective

*written by Andreas Horndahl*

In this section, we cover key aspects of leveraging annotated data for business applications. We discuss generating business value (Sec. 8.1), impact assessment (Sec. 8.2), and managing annotated data as a valuable asset (Sec. 8.3). These concepts guide businesses in maximizing the potential of annotated data while ensuring efficient resource allocation and data security.

## 8.1 Annotated Data and Business Value

The primary goal of annotating data is to create business value in some form. Business value can, for instance, be created by training a machine learning model that can be used to find out what your customers think about your products via sentiment analysis (e.g. Fig. 1) of product reviews. Business value can also be created by facilitating development of new capabilities, such as extracting key information elements from job ads which can be used to create multiple downstream applications.

Since an annotated dataset does not by itself provide business value, it is vital to ensure that all the components and processes needed for creating business value are taken into account when a cost-benefit analysis is done. A holistic approach must be taken that considers multiple factors, such as the ease of maintaining and improving an AI-based application as a whole, and it is essential to have a mindset that embraces the need for continuous improvements. This involves establishing a process for adding new training data, re-training the model, and deploying it to production. Version control for both the model and the datasets used for training the model is also necessary to be able to improve the model based on feedback and/or unexpected behavior.

As discussed in previous sections, annotating data is often a costly and time-consuming process. The cost of creating an annotated dataset depends on factors such as the size and complexity of the dataset, the number of annotators needed, and the rate of pay. It's important to understand the trade-offs between the cost of annotation and the potential benefits in terms of increased model performance and business value. For example, investing in creating a high-quality dataset may be more expensive initially, but can result in a more accurate model that delivers significant long-term benefits. However, it is important to keep in mind that model performance is not the same thing as business value. In some scenarios, an 80% accuracy might be sufficient and the work needed to improve the model performance above this point may not be worthwhile. Although the annotation work may be expensive in and of itself, the total cost includes other things such as obtaining appropriate annotation tools and setting up feedback loops for continuous improvements. Since the estimated total cost may be high, it may be worthwhile to take into account all potential business opportunities that can be developed based on the same dataset in the cost-benefit analysis.

## 8.2 Impact Assessment

Before beginning the data annotation process, and during its progression, it is important to understand the impact on the customer experience and business value. In practice, this means that the intended use case must be clearly understood and be taken into account when specifying *what* to annotate and *how* since the choice of labels and samples can significantly impact the accuracy and effectiveness of the model. The use case also dictates how to evaluate the model that is developed based on the annotated data.

The significance of the intended use case can be illustrated by the following scenario. A number of randomly selected job advertisements are annotated to create a model that can extract phrases that represent competencies from the texts (see examples in **Fig. 1**). Let's assume that the goal is to develop an application where a user can explore vacant positions from a competence perspective. With this application, users can gain a better understanding of the competencies typically expected for a given role, and can highlight those competencies in their job applications. One relevant question in this context is whether the model will be able to find all competencies for all professions since job advertisements for different professions often exhibit distinct styles and levels of information detail. If a category of professions is underrepresented among the examples used to train the model, there is a risk that the model performs less well on that subset. If the model performs poorly in extracting competencies related to certain professions, the user may get a false impression of the required competencies for some professions. In this case, it is advisable to examine if there is any imbalance in the dataset and to evaluate the model's performance for each profession separately. By doing so, you can identify any areas where the model may be underperforming and take corrective action, such as adjusting the annotation guidelines or adding more training data.

When a high-quality dataset has been developed, it may be tempting to reuse it for other use-cases. In some cases, the dataset can be reused as is but this is not always the case since the requirements may vary. Hence, it is important to understand how the dataset was annotated, the assumptions made during the annotation process, and the downstream application the annotators had in mind before reusing the dataset for another use-case. For example, if the use-case is to let users explore terms related to roles in vacation positions, it does not matter in practice if terms like "programming" are extracted as a hard skill or a job description by a model. However, if the use case is to compare different occupations based on distinct term categories (e.g. hard skills, soft skills, and task descriptions) to generate job recommendations based on similarity, it might be more problematic if terms are associated with an incorrect label. To reach the level of accuracy needed for the latter use case, it may be necessary to adapt the dataset by updating the annotation guideline, correcting some annotations, and creating a new model that is better suited for the new use-case.

## 8.3 The Value of Annotated Data

A significant amount of time and resources may have been invested in developing an annotated dataset. The dataset should therefore be treated with the same level of importance and care as any other critical business asset. In many cases, the annotated data is as valuable as the final model itself. Proper management includes implementing secure data management protocols, version control, backup procedures, and access controls, to ensure that the data is protected from unauthorized access or misuse.

Additionally, as annotated data may contain sensitive information, privacy protection measures are necessary to comply with data protection regulations and maintain the trust of individuals. To achieve this, encryption, access controls, and anonymization techniques may be necessary to ensure that sensitive information is not exposed or compromised.

# 9 Ethics and Regulations

*written by Andreas Horndahl*

The field of artificial intelligence (AI) has made substantial progress in recent years leading to disruptive advancement across various industries. As AI capabilities evolve and facilitate the creation of novel applications, it becomes increasingly important to address concerns related to trustworthiness and ethical implications.

How a machine learning model, or a complete AI system behaves from an ethical perspective is closely tied to the quality and nature of the data used for training the model. Thus, the annotation process, which involves labeling and categorizing data, plays an important part in ensuring that the model is good enough from an ethical perspective.

## 9.1 The Artificial Intelligence Act

The Artificial Intelligence Act (AI Act) is a regulation proposed by the European Commission (April 21, 2021) with the goal of creating a legal framework for the use of artificial intelligence [10]. The regulation targets individuals and organizations involved in creating and utilizing AI in their operations or activities. The regulation is planned to be adopted in 2023 and will take effect two years afterwards.

The proposed regulation defines AI as a range of software development frameworks that include machine learning, expert systems, and statistical approaches. The AI Act categorizes AI applications into four risk categories: unacceptable risk, high risk, limited risk and minimal risk. High-risk applications can be found in areas such as employment, credit and healthcare, where they have the potential to impact individuals or society. High-risk applications are subject to ambitious legal requirements related to the use of data, record-keeping, and the allocation of responsibilities among those involved in their development and use. For example, the training data sets for these systems must meet specific criteria regarding relevance, representation, and accuracy. This includes ensuring that training, validation, and test datasets are free of errors and as complete as possible. To meet these requirements, it is essential that the data annotation process is well thought-out and documented. While "free of errors" is not explicitly defined, it is an ambitious objective since data annotation work often involves trade-offs between data quality and the time, effort and cost required to achieve it. Even with the most rigorous annotation guidelines and quality control procedures, it's very difficult to create a dataset that is completely error-free. However, it is desirable to minimize errors to the best extent possible and to have clear guidelines and quality control procedures in place to identify and correct any errors that may occur. While limited-risk applications are not subject to the same rigorous regulations, they must comply with transparency guidelines. For instance, system providers need to ensure that AI systems designed to interact with people clearly communicate to users that they are interacting with an AI system.

## 9.2 Guidelines and Tools for Trust

Since AI ethics is becoming more relevant and regulations like the AI Act will soon be implemented, it is important to establish good practices to ensure that AI systems are developed in a way that fosters trust. This implies that organizations must be able to explain in detail how data that was used to develop machine learning (ML) models was collected and annotated. Organizations that utilize AI technology should also be able to describe how they ensure that the data used to train models is representative and how the strategies put in place address bias.

Fortunately, tools and guidelines are in development to aid organizations to develop trustworthy AI systems. For example, a so-called "trust model" has been developed as part of a government initiative to promote public administrations ability to use AI [11]. In practice, the trust model is a self-assessment tool consisting of a wide range of questions covering multiple aspects that an organization is expected to answer before and during the development of AI-based applications. The questions aim to raise awareness and enable informed decision-making in the design of AI solutions, including the acquisition and annotation of data. The primary aim of the tool is to increase openness and transparency within the Swedish public administration. For example, the tool can be used to create a standardized description of an AI-system for display to regulatory authorities and interested members of the public.

## 9.3 Bias and Discrimination

The quality and quantity of the data are crucial factors when developing a ML model. Any biases or weaknesses present in the training data may be passed on to the model and could potentially lead to systematic discrimination or unsatisfactory user experiences. It is especially important to pay attention to the characteristics of the dataset, and the model's performance, if the model is used for any type of automatic decision-making that could impact an individual's life.

According to the Swedish Equality Ombudsman (Diskrimineringsombudsmannen, DO) , deficiencies in data in the context of machine learning can, in the general case, pose a risk of discrimination when [12]:
A. Necessary data is missing
B. Data used does not reflect actual conditions and has been influenced by bias during collection
C. Data reflects historical discrimination and reproduces it
D. Data is interpreted in a way that leads to a risk of discrimination

In the context of annotation, there is a risk that those annotating the dataset contribute to discrimination through the choices and interpretations they make (D). Additionally, the selection of documents to be annotated can also contribute to discrimination if not done properly (see **Sec. 7.3**). To illustrate how discrimination may occur, imagine a scenario in which an organization seeks to automatically evaluate a large number of CVs and cover letters. Suppose that numerous CVs and cover letters are labeled according to their quality and relevance in relation to the organization's values and requirements. In this scenario, there are several factors that may lead to discrimination. Some annotators might unintentionally link foreign surnames to less favorable outcomes while other annotators could baselessly perceive the choice of words and phrasing as negative indicators.

The most suitable method to avoid bias and discrimination when annotating a dataset varies, depending on the particular dataset and application. However, here are some general guidelines on how to minimize bias and discrimination [13] [14]:

**Raising awareness**
The individuals annotating the raw data should be aware that they may contribute to bias or discrimination by the interpretations they make. One way to raise awareness is by highlighting potential pitfalls and providing illustrative examples of how data should be interpreted in the annotation guidelines (see **Sec. 3.2**).

**Narrowing the problem scope**
A clear and limited problem scope makes it easier to ensure that the model is performing well. Attempting to address too many problems at once can lead to a complicated class system (see **Sec. 7.2**), which consequently makes the annotation guidelines challenging to follow.

**Considering multiple opinions**
Involving a diverse group of individuals with varying characteristics, such as gender, age, experience, and culture, involved in the annotation process can lead to a wider range of perspectives, thereby increasing the likelihood of identifying potential issues (see **Sec. 4**).

**Understanding and owning your data**
By gaining a thorough understanding and ownership of the data, it becomes less likely to be caught off guard by problems such as imbalance in the dataset. It is important to ensure that the data represents the full diversity of end users.

**Feedback and adjustments**
Even if preventive measures are in place to maximize the likelihood that the model produces acceptable predictions, it is still possible that the model produces inappropriate output. It's important to keep in mind that avoiding discrimination is normally not covered by the loss function used to train the ML model. Since it's impossible to guarantee that a ML model produces acceptable predictions every time, it is good to be prepared to be able to act swiftly if problematic cases are discovered. Appropriate adjustment may involve updating the annotation guideline and reannotating data..

http://urn.kb.se/resolve?urn=urn:nbn:se:kth:diva-305901

# Appendix

## A: Machine Learning in a Nutshell

*written by Joey Öhman*

This section aims to give the reader a very basic understanding of machine learning and pre-trained language models. If the reader already has some knowledge of machine learning this section can be skipped.

## Supervised Learning

While the goal of annotation can vary, the most common reason for annotating data is to provide labels that can be used in Supervised Learning to train a machine learning model. A machine learning model typically takes an input-output pair and approximates a function that generates the output, given the input. Modern machine learning models usually take the form of neural networks that can theoretically approximate any function, given sufficient model size and data.

In practice, it is not necessary to understand how these neural networks learn from data. It is enough to understand the role of the inputs and outputs to see why annotation is often required. Moreover, there are existing libraries or frameworks that will handle the machine learning part so long as you have the data in place. The most common case is that there are raw data (inputs) available but no labels (outputs). Therefore, the work of achieving a good model often primarily consists of defining the problem and creating these labels.

Formally, a model takes inputs-output pairs **(x, y)** and learns a function **f**, that maps inputs to outputs: **f(x) = y**. The outputs **y** can be categorical (e.g. classes A, B) or numerical (e.g. numbers 1.3, 22.2). In the case of categorical outputs, we speak of a classification problem. An example of this type is sentiment classification. In the case of numerical outputs, we speak of a regression problem. An example of a regression problem is sentence similarity.

Furthermore, the dataset **(X, Y)** composed of data points (**x**, **y**), is typically separated into three subsets: train, validation, and test. The train set is used to train the model, i.e. to learn **f**. The validation set is used to find good architecture-specific hyperparameters (e.g. learning rate and neural network topology), and to monitor the training to help identify when training should be stopped (e.g. to prevent overfitting). Lastly, the test set that has not been seen by the model until after training, is used to evaluate the model and estimate how well it generalizes to unseen data.

It is good to keep in mind that machine learning solutions are often not very robust out of the box and can make incorrect predictions. Instead, they have the potential to automate large-scale tasks that would be manually infeasible or enhance decision-making when domain experts are looking at all data points in any case.

## Pre-trained Language Models

When working with textual tasks, it is almost always best to use a pre-trained language model (LM), which is typically a form of Transformer network [15]. There are many available models on HuggingFace[5], that have been trained on massive amounts of text to gain a general understanding of the language. So, instead of training a neural network from scratch, a pre-trained LM is used as a base model and task-specific data is used to continue training (fine-tuning) the model.

When the input is text and the output is a number or a category, a BERT [16] model, or similar, is usually the best choice. Today, there are many options for both English[6] and Swedish[7] [17]. Since the best available models are changing as the field progresses, we recommend doing some research prior to training to find the best model for the task and language at hand.

### Example: Fine-tuning BERT

Consider the example use case introduced in **Sec. 1**, where one wants to classify user review sentiments. After annotation, when a dataset with labels is in place, a model—for instance a pre-trained BERT model—can be trained to automatically execute the task. The data points are usually split into train/validation/test subsets. The model is fine-tuned using the train and validation sets and lastly evaluated with the test set. If they deem the model's performance sufficient, their task of flagging posts can be successfully automated. However, in practice, these model-flagged posts may have to be manually verified to ensure that no posts have been wrongfully removed.

## B: Annotation Guidelines - Template

*written by Emma Wallerö*

Here follows an example template of annotation guidelines.

| Description | Example |
| --- | --- |
| 1. A short introduction to the problem containing the most important information | Your task is to annotate a number of reviews between 2014-2018 as positive, negative or neutral based on the feeling and message conveyed. You will do this classification task using three different labels. |
| 2. A description of the possible labels that can be used. | *Labels for sentiment classification* |

---

[5] https://huggingface.co/
[6] https://huggingface.co/bert-base-cased
[7] https://huggingface.co/KB/bert-base-swedish-cased

|   |   |
|---|---|
|   | **POS**<br>A review conveying a positive sentiment or message |
|   | **NEG**<br>A review conveying a negative sentiment or message |
|   | **NEU**<br>A review not conveying neither positive nor negative sentiment or message |
|   | Reviews must be assigned exactly ONE label each. |
| 3. Some examples. This is less important if your task is quite straightforward. | |

| Review | Sentiment-tag |
|---|---|
| "It was even more efficient than I had imagined" | **POS** |
| "This was one of the worst microwaves I ever had to use!" | **NEG** |
| "The product did not reach my expectations" | **NEG** |
| "I bought this microwave two months ago" | **NEU** |
| 4. Instructions on how to handle ambiguous cases | ● If a review contains both positive and negative sentiment or is ambiguous, the review should be classified as neutral |

## Additional comments

- Consider giving instructions regarding how to navigate and use the annotation tool. The amount of detail depends on the tool used as well as the annotators. These instructions can be a part of or separate from your annotation guidelines.
- Make sure you mention whether a data unit can be assigned several labels or simply one label if you perform a classification task.

## Adjustments

This template is optimal for a classification task. If you are performing some type of NER, consider adding instructions and ambiguous cases regarding entity boundaries.

## C: Inter-Annotator Agreement

*written by Emma Wallerö*

There are several ways to measure inter-annotator agreement. One of the most common metrics is the Cohen's Kappa coefficient which can be used for comparing annotations between two annotators. The Kappa metric is useful for several reasons, one being the fact that apart from calculating observed agreement, it does take the agreement expected by chance into account, according to Sim and Wright 2005 [18]. A similar metric is Fleiss' Kappa coefficient. This is a good option if you want to get one metric score for the agreement between more than two annotators.

When using a Kappa metric in order to evaluate inter-annotator agreement it is important to be aware of its arbitrary properties. It is hard to say where to draw the line between a bad and a good Kappa value. This depends on your task and its conditions. The Kappa coefficient spans from -1 up to 1. It does however generally fall between 0 and 1 [18]. Having a Kappa value close to 1 corresponds to almost perfect agreement. Using the Kappa metric will give you an indication of the quality of your data as well as annotation guidelines.

Another suitable metric is the f1 score. This is especially useful when measuring inter-annotator agreement for named entity recognition where a Kappa value will be unfairly increased due to the uneven distribution of annotated versus un-annotated data [19]. The f1 score does not take expected chance agreement into account, however. The f1 score is a harmonic mean between recall and precision, and when measuring the inter-annotator agreement between two annotators you assume one of the two labeled data-sets to be a gold standard (correct) and the other to be predictions [20].